\documentclass[runningheads]{llncs}
\usepackage{graphicx}
\usepackage{times}
\usepackage{makecell}
\usepackage{bm}
\usepackage{float}
\usepackage{epsfig}
\usepackage{bbding}
\usepackage{amsmath}
\usepackage{amssymb}
\usepackage{color}
\usepackage{url}
\usepackage[breaklinks,colorlinks,linkcolor=blue,citecolor=blue,urlcolor=blue]{hyperref}

\begin{document}
	%
	\title{DUFormer: Solving Power Line Detection Task in Aerial Images using Semantic Segmentation}
	\titlerunning{DUFormer}
	\author{Deyu An\inst{1,2,3,4,\star} \and
		Qiang Zhang\inst{1(}\Envelope\inst{)} \and
		Jianshu Chao\inst{2,3,4} \and
		Ting Li\inst{1,2,3,\star} \and
		Feng Qiao\inst{5} \and
		Zhenpeng Bian\inst{1} \and
		Yong Deng\inst{1}}
	\authorrunning{D. An et al.}
	%
	\institute{Autel Robotics, Shenzhen, 518000, China \\
		\email{zq18487102396@gmail.com} \and		
		Fujian College, University of Chinese Academy of Sciences, Fuzhou, 35000, China \and
		Quanzhou Institute of Equipment Manufacturing, Fujian Institute of Research on the Structure of Matter, Chinese Academy of Sciences, Quanzhou, 36200, China \and
		Fujian Agriculture and Forestry University, Fuzhou, 35000, China \and
		RWTH Aachen University, Aachen, 52062, Germany
	}
	%
	%
	%
	%
	\maketitle              
	\footnote{$\star$ Interns at Autel Robotics, Deyu An and Qiang Zhang contribute equally.\\ This work was partially supported by Guiding Project of Fujian Science and Technology Program (No. 2022H0042).}
	\begin{abstract}
		Unmanned aerial vehicles (UAVs) are frequently used for inspecting power lines and capturing high-resolution aerial images. However, detecting power lines in aerial images is difficult, as the foreground data (i.e., power lines) is small and the background information is abundant. To tackle this problem, we introduce DUFormer, a semantic segmentation algorithm explicitly designed to detect power lines in aerial images. We presuppose that it is advantageous to train an efficient Transformer model with sufficient feature extraction using a convolutional neural network (CNN) with a strong inductive bias. With this goal in mind, we introduce a heavy token encoder that performs overlapping feature remodeling and tokenization. The encoder comprises a pyramid CNN feature extraction module and a power line feature enhancement module. After successful local feature extraction for power lines, feature fusion is conducted. Then, the Transformer block is used for global modeling. The final segmentation result is achieved by amalgamating local and global features in the decode head. Moreover, we demonstrate the importance of the joint multi-weight loss function in power line segmentation. Our experimental results show that our proposed method outperforms all state-of-the-art methods in power line segmentation on the publicly accessible TTPLA dataset.
		
	\end{abstract}
	\section{Introduction}
	Unmanned aerial vehicles (UAVs) are more popular in applications such as geographic information systems (GIS), power line inspection, security surveillance, and agricultural and forestry protection, surpassing traditional tools and human labour. However, detecting power lines is challenging because of the slender characteristics of the power line and complex backgrounds. To address this problem, DUFormer, a CNN-Transformer hybrid algorithm, is specifically designed to detect power lines in aerial images.
	
\begin{figure}[h]
		\setlength{\abovecaptionskip}{-1.em}
		\begin{center}
			\includegraphics[width=0.70\linewidth]{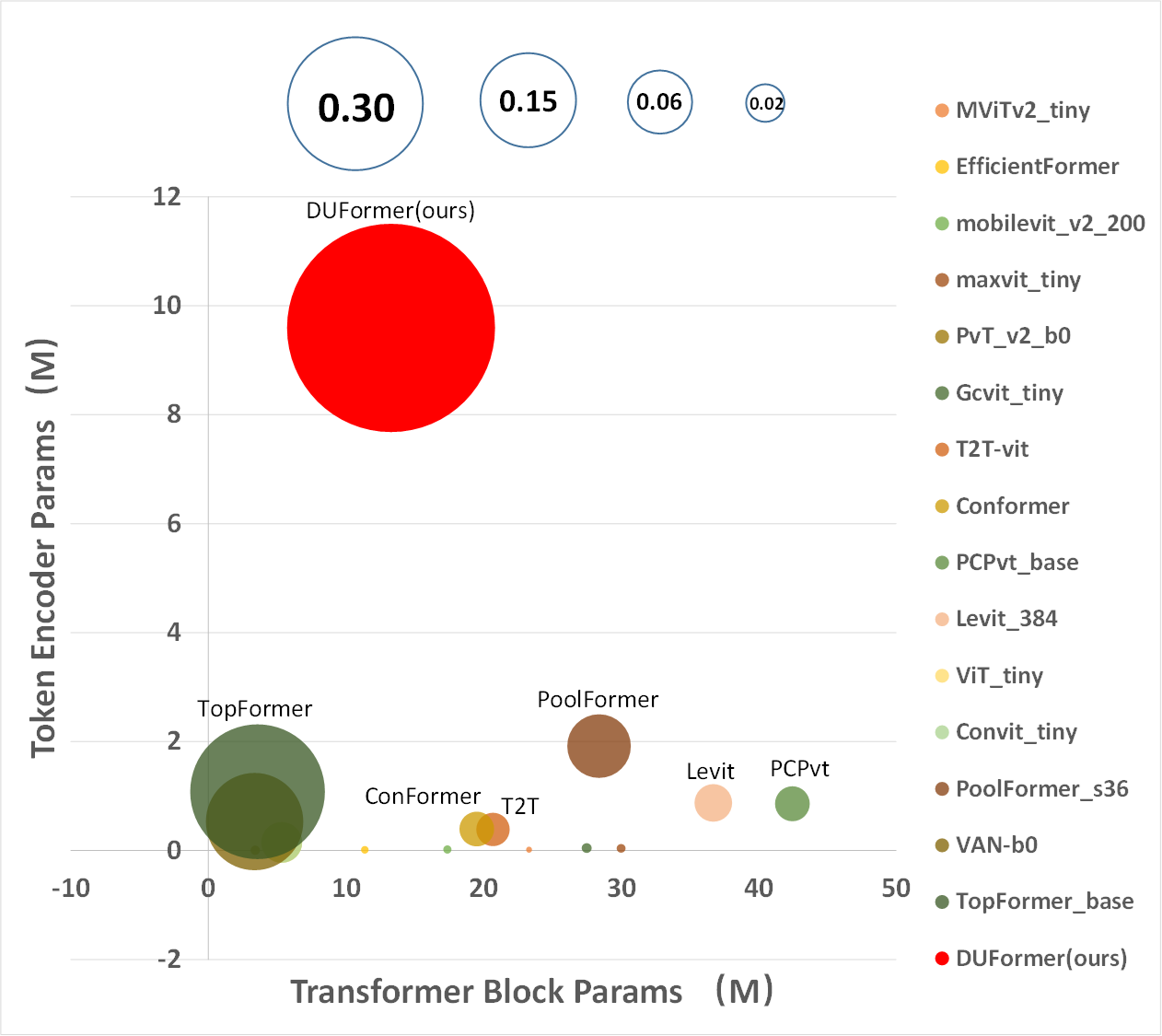}
		\end{center}
		\caption{\textbf{Parameter ratios}. The parameter ratio refers to the ratio of the token encoder's parameters to the Transformer blocks' parameters. The circle's size represents the parameter ratio.}
		\label{fig:params_ratios}
	\end{figure}

	Recently, the Vision Transformer~\cite{dosovitskiy2021an} and its variants have been used for challenging prediction tasks by performing global self-attention on high-resolution tokens. However, this approach is computationally and memory-wise inefficient as it results in quadratic complexity. Inspired by the work of Zhang et al. on TopFormer ~\cite{zhang2022topformer}, we argue that using a convolutional neural network (CNN) with a strong inductive bias to perform ample feature extraction could facilitate the training of an efficient Transformer model. Therefore, we first introduce the concept of a heavy token encoder. In the previous Transformer algorithms, as shown in Fig.~\ref{fig:params_ratios}, we observe that the parameter ratios of the token encoder (i.e., patch embedding) to the Transformer blocks are small, even less than 0.05. In contrast, the proposed DUFormer improves this parameter ratio to 0.7 with a heavy token encoder. The explanation of the heavy token encoder is elaborated in the upcoming sections.\vspace{-2.em}
	\vspace{-1.5em}
		
	\noindent
	We improve the efficiency of the Transformer model by leveraging the advantages of CNNs for feature extraction while maintaining the effectiveness of the Transformer's self-attention mechanism. First, the input feature maps are projected to four scales via a pyramidal CNN feature extraction module (DUB in Sec.~\ref{sec:dub}). Subsequently, the feature maps go through a transition layer to enhance the receptive field. Altogether, five feature maps with different scales are generated separately. Then they are tokenized by multi-scale average pooling and fed into the Transformer blocks for global attention calculation. Our method can produce tokens with relatively low resolutions and enables the Vision Transformer to perform computations with acceptable throughput, even when dealing with numerous feature map channels.
	
	In the local feature extraction stage of the power line detection task, we propose a power line feature enhance module, consisting of an asymmetric dilated convolution-based Power Line Aware Block (PLAB in Sec.~\ref{plab}) and a BiscSE module (in Sec.~\ref{sbiscse}). These modules extract slender power line features at shallow layers and enhance semantic information at deeper layers. The network follows the U-Net~\cite{ronneberger2015u} architecture, with the output of the Transformer block serving as the upsampling source. This output is upsampled four times in separate channels and then multiplied with our proposed power line aware block element-wisely before being concatenated with the DUB output in the corresponding decoding stage. Five stages generate five segmentation results, with losses calculated separately. The final segmentation result is obtained by fusing the five results.
	
	In aerial image power line detection tasks, imbalanced data samples pose a significant problem, as foreground (i.e. power line) pixels are considerably smaller than background pixels. To solve this, we introduce a joint multi-weight loss function. In our experiments, the proposed method outperforms existing methods, achieving state-of-the-art performance.
	
	In summary, our paper makes the following contributions:
	
	$\bullet$ We first propose the theory of a heavy token encoder and demonstrate that a Transformer model with sufficient token encoding is easier to train efficiently. Accordingly, we further propose a CNN-Transformer hybrid architecture for aerial image power line detection.
	
	$\bullet$ We propose a power line aware block for detecting slender power line features and an improved scSE module for enhancing the semantic information of the network at deeper layers.
	
	$\bullet$ We investigate the importance of the joint multi-weight loss, which improves the performance of the power line segmentation significantly.
	
	$\bullet$ Our approach achieves state-of-the-art performance on the TTPLA dataset by conducting a number of experiments.
	
	\section{Related Work}
	\subsection{Vision Transformer}
	
	The original Vision Transformer~\cite{dosovitskiy2021an} slices the image into multiple non-overlapping patches, which is good at capturing long-distance dependencies between patches but ignores local feature extraction. TNT~\cite{han2021transformer} further divides patches into multiple sub-patches and introduces a new structure, Transformer-iN-Transformer, which uses internal Transformer blocks to model the relationship between patches and external Transformer blocks to achieve patch-level information exchange. Twins~\cite{chu2021twins} and CAT~\cite{lin2022cat} alternate local and global attention layer by layer. Swin Transformer~\cite{liu2021swin} performs local attention in the window and introduces a shift window partitioning method for cross-window connections. In addition, some works combine CNN with the Transformer. CPVT~\cite{chu2021conditional} proposes a conditional position encoding (CPE) method, which is conditional on the local neighborhood of the input tokens. It applies to arbitrary input sizes for fine feature encoding using convolution. CVT~\cite{wu2021cvt}, CeiT~\cite{yuan2021incorporating}, LocalViT~\cite{li2021localvit}, and CMT~\cite{guo2022cmt} analyze the potential pitfalls of directly applying Transformer architecture to images. The mitigation method is proposed in their papers, i.e., combining convolution with Transformer. Specifically, a feedforward network (FFN) in each converter block is combined with a convolutional layer to facilitate the association between adjacent tokens.
	\subsection{Semantic Segmentation}
	The FCN~\cite{long2015fully} proposed by Long et al. in 2015 pioneered semantic segmentation in deep learning. It replaces the fully-connected layer in traditional CNN models with a convolutional layer, gradually deconvoluting to restore the original image size and obtain the final semantic segmentation result. In the same year, Ronneberger et al. proposed U-Net~\cite{ronneberger2015u}, also based on the FCN. The U-Net structure resembles the letter U with a encoding and decoding structure. Initially used for medical images, it is suitable for small datasets. In 2017, SegNet~\cite{badrinarayanan2017segnet}, proposed by Badrinarayanan et al., also has an encoder-decoder structure but with the difference that max-pooling with returned coordinates is used. Then, the returned coordinates are used for feature recovery during upsampling. As a result, it significantly reduces the model's parameters. PSPNet~\cite{zhao2017pyramid}, also proposed in 2017, introduced the pyramid pooling module (PPM). It concatenates four global pooling layers of different sizes to generate feature maps at different levels, aggregating multi-scale image features.
	The DeepLab series~\cite{chen14semantic,chen2017deeplab,chen2017rethinking,chen2018encoder} use dilated convolution to propose Atrous Spatial Pyramid Pooling (ASPP) and incorporate conditional random fields (CRF) in the final structured prediction to improve model accuracy.  Guo et al. proposed SegNeXt~\cite{guo2022segnext}, which introduces a new multi-scale convolutional attention (MSCA) module using a larger kernel size to capture global features. They designed the parallelization of multiple kernels to increase information combination of dense contexts, which is essential for semantic segmentation.
	\section{Proposed Architecture\label{Architecture}}
	\subsection{Overview}
	\begin{figure}[h]
		\setlength{\abovecaptionskip}{-1.em}
		\begin{center}
			\includegraphics[width=0.95\linewidth]{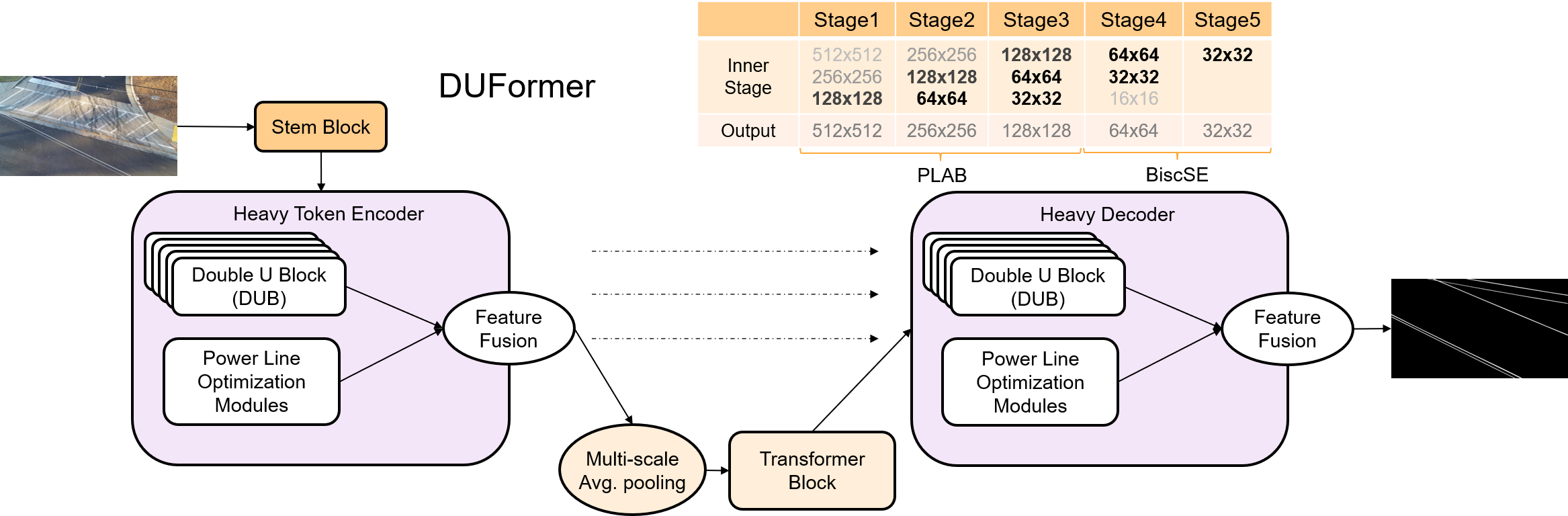}
		\end{center}
		\caption{Overall architecture of \textbf{DUFormer}. The table shows the feature map resolutions in different stages. The overlapping features in various stages represent our novel idea of feature re-mining.}
		\label{fig:overall}
	\end{figure}
	\noindent Our network for the high-resolution (1K) aerial image power line detection is presented in Fig.~\ref{fig:overall}. The overall architecture design follows the U-Net structure. The Stem block is designed to handle high-resolution data without consuming too much GPU memory or significantly increasing FLOPs. It consists of a parallel max-pooling layer and average-pooling layer, followed by channel feature fusion using a $1\times1$ convolution. The heavy token encoder implementation is divided into two parts: the pyramidal Double U Blocks (DUB in Sec.~\ref{sec:dub}) for obtaining feature maps with different resolutions and the Power Line Optimization Module, including PLAB (Sec.~\ref{plab}) and BiscSE (Sec.~\ref{sbiscse}) for enhancing power line features. Then, the output results of the DUB and Power Line Optimization Modules are fused. The fused result is used as the output of the heavy token encoder, and the feature fusion method is shown in Equation~\ref{stage123} and Equation~\ref{stage45}:
	\begin{equation}
		\begin{split}
			output_{stage_{1,2,3}}= out_{DUB_{1,2,3}} \odot out_{PLAB_{1,2,3}}\label{stage123}
		\end{split}
	\end{equation}
	\begin{equation}
		\begin{split}
			output_{stage_{4,5}}= out_{DUB_{4,5}} \odot out_{BiscSE_{4,5}}\label{stage45}
		\end{split}
	\end{equation}
	The Transformer block can effectively capture information about the thin power lines in the entire image. Its input needs to be tokenized, and the operation of tokenization is shown in Equation~\ref{transin}:
	\begin{equation}
		\begin{split}
			x = concat(A&daptiveAvgPool(output_{stage_{1}}),\\&...,AdaptiveAvgPool(output_{stage_{5}})
			\label{transin}
		\end{split}
	\end{equation}
	In the Transformer block, a multi-headed attention mechanism is used for global modeling. The multi-head attention is formulated as follows:
	\begin{equation}
		\begin{split}
			MultiHead(Q,K,V) = Concat(head_1,\dots,head_h) \cdot W^O
		\end{split}
	\end{equation}
	where $W^O$ is the learnable weight that maps the concatenate result back to the input dimension, $head_i$ represents the calculation of the $i^{th}$ attention head, as shown in Equation~\ref{headi}:
	\begin{equation}
		\begin{split}
			Head\_Attention_{i}(Q_{i},K_{i},V_{i})&=softmax\left(\frac{Q_{i}K_{i}^T}{\sqrt{d_{k_{i}}}}\right)V_{i} 
			\label{headi}
		\end{split}
	\end{equation}
	where the $softmax$ function is applied to the rows of the similarity matrix and $d_{k_{i}}$ provides a normalization. $Q_{i}$, $K_{i}$, $V_{i}$ represent the query matrix, key matrix, and value matrix, respectively.
	
	Finally, The decoder part is symmetric to the encoder in its structure. Following sections provide detailed descriptions of our proposed components.
	
	\subsection{Double U Block (DUB)\label{sec:dub}}
	\begin{figure}[h!]
		\setlength{\abovecaptionskip}{-1.em}
		\begin{center}
			\includegraphics[width=0.65\linewidth]{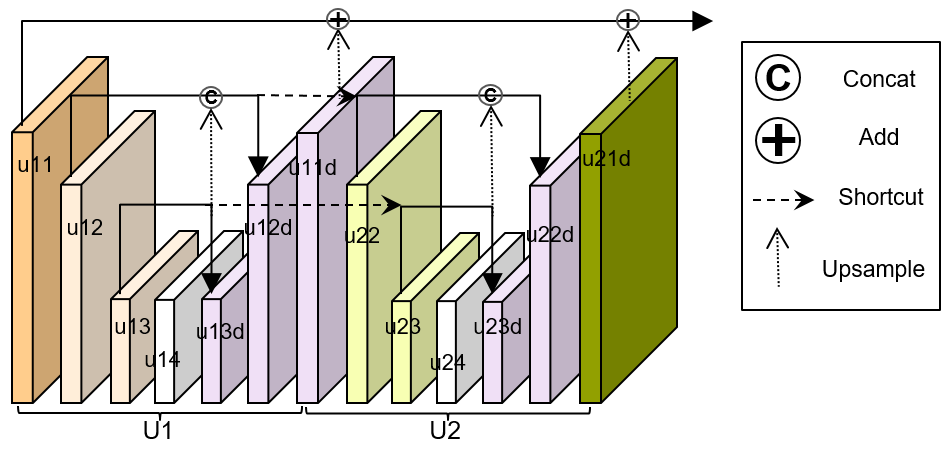}
		\end{center}
		\caption{Architecture of \textbf{D}ouble \textbf{U}-\textbf{B}lock.} \label{fig:dub}\vspace{-1.5em}
	\end{figure}
	\noindent
	\textbf{D}ouble \textbf{U} \textbf{B}lock (DUB) is an integral part of the heavy token encoder and comprises two U-shaped networks named U1 and U2, as illustrated in Fig.~\ref{fig:dub}. Each U-shaped network consists of two downsampling and upsampling layers, resulting in a larger receptive field. Furthermore, the feature maps of each resolution in U1 and U2 are connected through a shortcut, which allows information not mined in U1 to be mined again in U2. This method enhances the network's information mining capability. The joint output of residuals from both shallow and deep features is beneficial to constructing deep networks.
	
	The feature extraction component of DUFormer consists of four DUBs and one transition layer. Each DUB operates at a specific multi-level resolution as shown in Fig.~\ref{fig:dub} and employs overlapping feature mining, as shown in the table located in the upper-right corner of Fig.~\ref{fig:overall}. The degree of repetition in feature map mining is indicated by the darkness of its corresponding color. The transition layer employs dilated convolution, capturing features at different scales without downsampling or upsampling the feature maps. In other words, big receptive field ensures the capture of fine-grained detail while maintaining a high-level understanding of the input image. This method facilites effective global modeling by the subsequent Transformer block.

	\subsection{Power Line Aware Block (PLAB)\label{plab}}
	\begin{figure}[h]
		\setlength{\abovecaptionskip}{-0.5em}
		\begin{center}
			\includegraphics[width=0.60\linewidth]{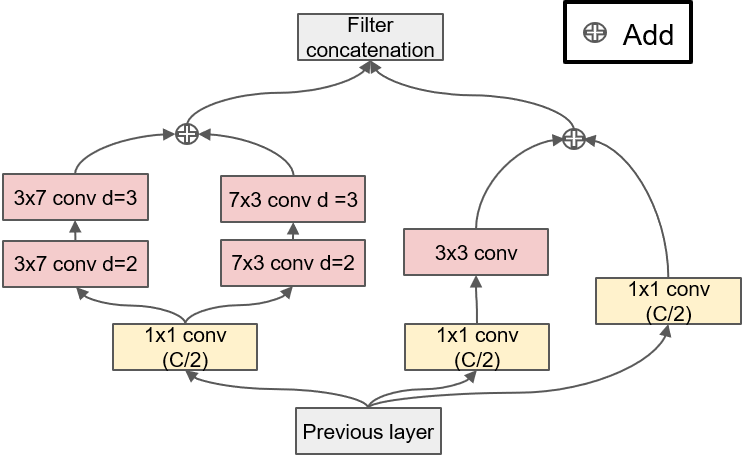}
		\end{center}
		\caption{Architecture of \textbf{P}ower\textbf{L}ine \textbf{A}ware \textbf{B}lock.  }
		\label{fig:plab}\vspace{-1.5em}
	\end{figure}
	\noindent
	We present the PLAB, a module designed to effectively extract slender power line features from high-resolution aerial images. The module is designed to leverage the rich, detailed information available in the shallow network layers to extract power line features precisely.  The PLAB structure, illustrated in Fig.~\ref{fig:plab}, comprises two parallel asymmetric dilated convolutions that efficiently extract features in both vertical and horizontal directions, while also complementing the features with paralleled original convolutions. The feature fusion enables the subsequent network to exhibit an enhanced ability to perceive power lines.

	\subsection{BiscSE Block\label{sbiscse}}
	\begin{figure}[h]
		\setlength{\abovecaptionskip}{-0.5em}
		\begin{center}
			\includegraphics[width=0.60\linewidth]{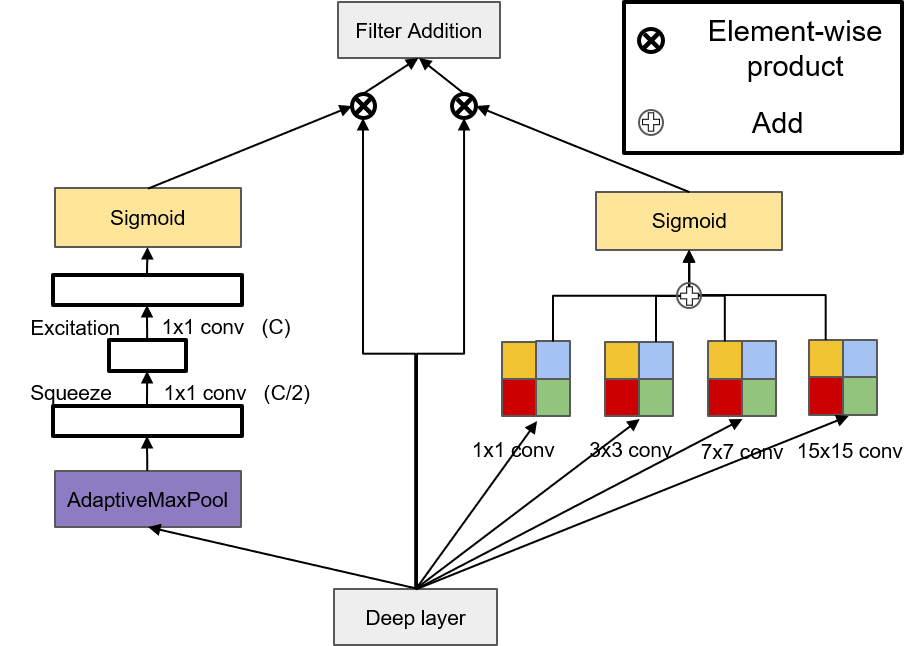}
		\end{center}
		\caption{Architecture of \textbf{BiscSE} }
		\label{fig:Biscse}\vspace{-1.5em}
	\end{figure}
	\noindent
	Although the feature map resolution is lower in the network's deep layers, the extracted semantic features are more robust. To enhance the effectiveness of the scSE~\cite{roy2018recalibrating} module as depicted in Fig.~\ref{fig:Biscse}, we replace the average-pooling with max-pooling in the channel SE branch. Max pooling retains the maximum activation within each pooling region, which helps capture the most discriminative features (i.e., power line feature) present in the input data. This is particularly beneficial for power line detection tasks as it enhances the localization of power line feture in channel dimension. In addition, we extend the original module to improve the spacial SE branch by using various convolutional kernels for feature extraction under different receptive field. This idea achievs proper spatial squeeze and expansion, and further enhances the performance of the spatial SE branch.
	
	\subsection{Loss Function}
	In conventional semantic segmentation algorithms, the input images typically includes valid semantic information. However, in the power line detection task, the background can introduce much irrelevant and redundant information. Using the cross-entropy loss function, which is a standard loss function for semantic segmentation, may not adequately suppress this redundant information. To address this issue, we investigate multiple loss functions, i.e., FocalLoss, PhiLoss, and DiceLoss, and devise an approach to combine them to improve the model's focus on detecting power lines. The combination is formulated as follows:
	\begin{equation}
		\begin{split}
			FocalLoss(p,\hat{p}) &= -(\alpha(1-\hat{p})^{\gamma} plog(\hat{p})\\
			&+(1-\alpha)\hat{p}^\gamma(1-p)log(1-\hat{p}))\label{focal}
		\end{split}
	\end{equation}
	\begin{equation}
		PhiLoss = (1-MCC)^{\theta}\label{philoss}
	\end{equation}
	\begin{equation}
		DiceLoss = 1-{\frac{2*TP}{2*TP+FP+FN}}	
	\end{equation}
	\noindent
	where $\alpha$ in Equation ~\ref{focal} is used to adjust the ratio between positive and negative sample losses, $\gamma$ is used to reduce the loss contribution of the easy samples. $MCC$ in Equation~\ref{philoss} refers to the Matthews correlation coefficient. Equation~\ref{multi} shows the final loss used in our experiments. By conducting the experiment, we get the values of $\rho$, $\tau$, $\phi$ as 3.0, 1.5, 3.0, respectively.
	\begin{equation}
		\begin{split}
			Loss=&\rho*FocalLoss_{weight=1:5}+\tau*PhiLoss\\&+\phi*DiceLoss_{weight=1:5}\label{multi}
		\end{split}
	\end{equation}
	
	\section{Experiments}
	In this section, we evaluate DUFormer's ability to detect power lines by conducting many comparative and ablation experiments and demonstrate the reliability and feasibility of our proposed model. We compare the proposed method with classical and state-of-the-art algorithms.
	\subsection{Experimental Settings}
	\subsubsection{Datasets.} We conduct experiments on the challenging power line dataset TTPLA~\cite{abdelfattah2020ttpla}, which consists of 1124 training images and 107 validation images with a resolution of $3840\times2160$. 
	\vspace{-1.0em}
	\subsubsection{Training.} Our proposed approach is implemented based on the MMSegmentation framework. The network model is trained from scratch, specifically for power line data, without using any pre-trained weights. We set the training maximum iteration to 80k, the initial learning rate to 9e-4, and the weight decay to 0.01 for all experiments. We use a 'poly' learning rate strategy with a factor of 1.0. To ensure fair comparisons, we fix the random seeds in all experiments. A batch size of 8 is used, and all experiments are conducted on two NVIDIA GeForce RTX 3090 GPUs.
	\vspace{-1.0em}
	\subsubsection{Testing.}We test with an inference resolution of $1080\times1920$ for all the methods during the test procedure.
	\vspace{-1.0em}
	\subsubsection{Evaluation criteria.}When detecting power lines in the UAV aerial data, we aim to achieve a model with high sensitivity, which reflects in a high Recall value, i.e., a low power line miss-detect rate. However, we must also consider Precision and ensure it falls within an acceptable range while maintaining a high Recall score. Therefore, we adjust the calculation of the F-score accordingly.
	\begin{equation}
		\operatorname{F-score}=(1+\beta^2)*\frac{Precision*Recall}{\beta^2*Precision+Recall}
	\end{equation}
	where $\beta$ is set to 2, giving more weight to Recall than Precision and making the F-score consistent with our desired low power line miss-detect rate.
	\subsection{Comparative Experiments}
	\subsubsection{DUFormer vs. other methods.}
	Table~\ref{comparision} presents the performance of our DUFormer method for power line detection, demonstrating superior results compared to other classical methods. Our method achieves an F-score of 85.96\% and an IoU of 74.4\% with only 28.51M parameters, outperforming HRNet-OCR~\cite{YuanCW20}, which has 70.37M parameters. DUFormer only accounts for 1/5 of HRNet-OCR's FLOPs, which is a substantial improvement in limited computing resources. Furthermore, we test the inference speed of each model, The proposed method outperforms other models with $200.40 ms/img$, which can effectively improve the efficiency of power line detection tasks.
	\begin{table}[h]
		\begin{center}
			\caption{Comparison results of various models.}\label{comparision}
			\begin{tabular}{l|l|l|l|l|l|l|l}
				\hline
				Method & \makecell{\#Params\\(M)} & \makecell{FLOPs\\(G)} &  \makecell{F-score\\(\%)} & \makecell{Precision\\(\%)} & \makecell{Recall\\(\%)} & \makecell{IoU\\(\%)} & \makecell{Latency\\(ms)} \\
				\hline
				Deeplab & 29.06 & 813.71 & 81.03 & 85.61 & 79.96 & 70.48 & 763.35\\
				PSPNet & 29.05 & 791.02 & 81.16 & 85.57 & 80.13 & 70.59 & 833.33 \\
				Sem-FPN-r50 & 28.51 & 182.48 & 80.03 & 84.17 & 79.06 & 68.82 &366.30 \\
				U-Net & 29.06 & 810.23 & 82.11 &$\bm{86.95}$ & 80.99 & 72.21 & 401.61 \\
				SegNeXt-Base & 28.0 & 128.1 & 79.07 & 82.26 & 78.31 & 67.17 & \underline{361.01}\\
				SegFormer-b2 & 24.76 & 74.21 & 82.51 & 85.05 & 81.9 & 71.59 & 543.47 \\
				EncNet & 35.89 & 563.28 & 76.14 & 82.08 & 74.78 & 64.28 & 520.8 \\
				CCNet& 49.81 & 801.52 & 78.04 & 83.48 & 76.79 & 66.66 & 595.23\\
				HRNet-OCR& 70.37 & 648.39 & \underline{83.91} & 86.0 & \underline{83.41} & \underline{73.43} & 800.00\\
				DUFormer(ours) & 28.51 & 123.41 & $\bm{85.96}$ & 84.35 & $\bm{86.37}$ & $\bm{74.44}$& $\bm{200.40}$\\
				\hline
			\end{tabular}	
		\end{center}
	\end{table}
	\vspace{-2.5em}
	\subsubsection{Joint multi-weighted loss function.}  In the following experiments, we investigate the effectiveness of the joint multi-weight loss function in addressing category imbalance issues in power line detection due to the slender characteristics of the power line. We employ the traditional Cross-Entropy loss function as a baseline and demonstrate in Table~\ref{loss} that using the joint multi-weight loss function improves the network's ability to detect power lines and reduces cluttered background information. Moreover, we apply the joint multi-weight loss function to other existing algorithms, e.g., SegNeXt~\cite{guo2022segnext}, EncNet~\cite{Zhang_2018_CVPR}, and CCNet~\cite{huang2018ccnet}. We observe a significant improvement in the Recall metrics, indicating a lower power line miss-detect rate. These similar observations further validate the effectiveness of our proposed method.
	\begin{table}[h]
		\begin{center}
			\caption{Results of  different loss functions.}\label{loss}
			\begin{tabular}{l|l|l|l|l|l}
				\hline
				Method & Loss & Fscore(\%)&Precision(\%) & Recall(\%) & IoU(\%) \\
				\hline
				DUFormer (ours) & CE Loss &82.38& 86.44  & 81.42 & 72.2 \\
				DUFormer (ours) & Multi Loss &$\bm{85.96}$ & 84.35 & $\bm{86.37}$ & $\bm{74.44}$\\
				\hline
				SegNeXt &CE Loss&  79.07 & 82.26 & 78.31 & 67.17 \\
				SegNeXt & Multi Loss & $\bm{81.91}$& 78.35 & $\bm{82.85}$ & $\bm{67.49}$  \\
				\hline
				EncNet & CE Loss &76.14&  82.08 & 74.78 & 64.28 \\
				EncNet & Multi Loss &$\bm{82.67}$ & 79.48 & $\bm{83.51}$ & $\bm{68.7}$\\
				\hline
				CCNet & CE Loss &78.04&  83.48 & 76.79 & 66.66 \\
				CCNet & Multi Loss &$\bm{82.73}$ & 78.6 & $\bm{82.83}$ & $\bm{68.26}$\\
				\hline
			\end{tabular}
		\end{center}
	\end{table}

	\subsection{Ablation Experiments}
	\subsubsection{Effect of heavy token encoder.} As noted in the previous section, the heavy token encoder can significantly enhance the Transformer models' performance on small data sets. Table~\ref{heavy} presents the experimental results. The lightweight token encoder serves as the baseline, and the repetitions of the Transformer block produce minimal improvement. In contrast, when the proposed heavy token encoder is applied, the performance improves significantly, resulting in a 3.2\% increase in both IoU and F-score compared to the baseline.
	\begin{table}
		\begin{center}\vspace{-1.5em}
			\caption{Results of Heavy Token Encoder.}\label{heavy}
			\begin{tabular}{l|l|l|l|l|l|l|l}
				\hline
				Method & \makecell{Tok. Encoder\\Parameters} & \makecell{Trans. Block\\Parameters} & \makecell{Param.\\Ratios} & \makecell{F-score\\(\%)}  & \makecell{Precision\\(\%)} & \makecell{Recall\\(\%)} & \makecell{IoU\\(\%)} \\
				\hline
				Baseline & 0.881 & 13.28 & 0.066 & 82.76 & 83.96 & 82.46 & 71.24 \\
				\hline
				\makecell[l]{Repetitive \\ Transformer block} & 0.881 & 26.56 & 0.033 & 84.09 & 82.25 & 84.56 & 71.51 \\
				\hline
				\makecell[l]{Heavy Token \\ Encoder (ours)} & 9.594 & 13.28 & 0.722 & $\bm{85.96}$ & $\bm{84.35}$ & $\bm{86.37}$ & $\bm{74.44}$\\
				\hline
			\end{tabular}
		\end{center}
	\end{table}\vspace{-1.9em}

	\subsubsection{The impact of each module.} Table~\ref{each} shows the impact of each module on the model. First, we establish a baseline model by removing the proposed modules from the network. 
	
	\begin{table}[h]
		\begin{center}\vspace{-1.5em}
			\caption{The impact of each module.}\label{each}
			\begin{tabular}{l|l|l|l|l|l|l}
				\hline
				+DUB & \makecell{+PLAB\&BiscSE} & +Multi Loss & \makecell{Fscore(\%)} & \makecell{Precision(\%)} & \makecell{Recall(\%)} & \makecell{IoU(\%)} \\
				\hline
				\color[RGB]{128,128,128}{\XSolidBrush}&\color[RGB]{128,128,128}{\XSolidBrush}&\color[RGB]{128,128,128}{\XSolidBrush} &81.01 &81.11&80.98&68.14  \\
				\CheckmarkBold&\color[RGB]{128,128,128}{\XSolidBrush}&\color[RGB]{128,128,128}{\XSolidBrush}&82.13 & 84.29     &81.6 &70.83\\
				\CheckmarkBold&\CheckmarkBold&\color[RGB]{128,128,128}{\XSolidBrush}&82.38&86.44&81.42& 72.2\\
				\CheckmarkBold&\CheckmarkBold&\CheckmarkBold  &$\bm{85.96}$&84.35&$\bm{86.37}$ &$\bm{74.44}$\\ 
				\hline
			\end{tabular}
		\end{center}
	\end{table}
	\noindent
	Then, we gradually add each proposed module to show their contributions. The results show that Double U Block, Power Line Aware Block and BiscSE can significantly improve the Precision. It should be noted that the high Recall is more in line with the industrial requirements, as the power lines in the aerial images should be detected in their entirety as much as possible. The Multi-Loss function is able to regulate Recall and Precision better, making the model more sensitive (i.e., higher Recall) to power lines. In summary, our proposed methods can enhance the power line detection capability of the model in various aspects.
	We further evaluate the performance of the PLAB at different positions in the network and compare the effectiveness of the improved scSE module. The experimental results can be found in Section 2 in the supplementary material. The supplementary material can be found at  \url{https://drive.google.com/file/d/194qGtaBfF9zuCGXuxSZ8uax1hZH0Sc9x/view?usp=sharing}.

	\section{Conclusion}
	In this work, we present a comprehensive methodology for tackling the problem of detecting power lines in aerial images captured by UAVs. The proposed solutions include a heavy token encoder, which captures fine-grained features by performing feature re-mining on feature maps of different resolutions at different stages. We also introduce a Power Line Aware Block (PLAB), composed of asymmetric dilated convolutions, which particularly enhances power line features while suppressing background information. Moreover, we propose an improved scSE module, i.e., BiscSE, optimized for dichotomous segmentation for power lines, effectively enhancing the Precision and Recall metrics. Through extensive experiments, we demonstrate that our proposed method significantly improves the performance in terms of the number of parameters, computational complexity, and accuracy. The proposed DUFormer sets a new state-of-the-art record on the public dataset TTPLA.
	
	%
	%
	%
	\bibliographystyle{splncs04}
	\bibliography{egbib}

\end{document}


\title{Supplemntary material\\DUFormer: Solving Power Line Detection Task in Aerial Images using Semantic Segmentation}
\titlerunning{DUFormer}
\author{}
\institute{}
\authorrunning{D. An et al.}
%

%
%
%
%
\maketitle

\vspace{-1.0em}
\section{Comparative Experiments}
\subsection{FLOPs, IoU and Parameters}
\begin{figure}
	\begin{center}
		\includegraphics[width=0.9\linewidth]{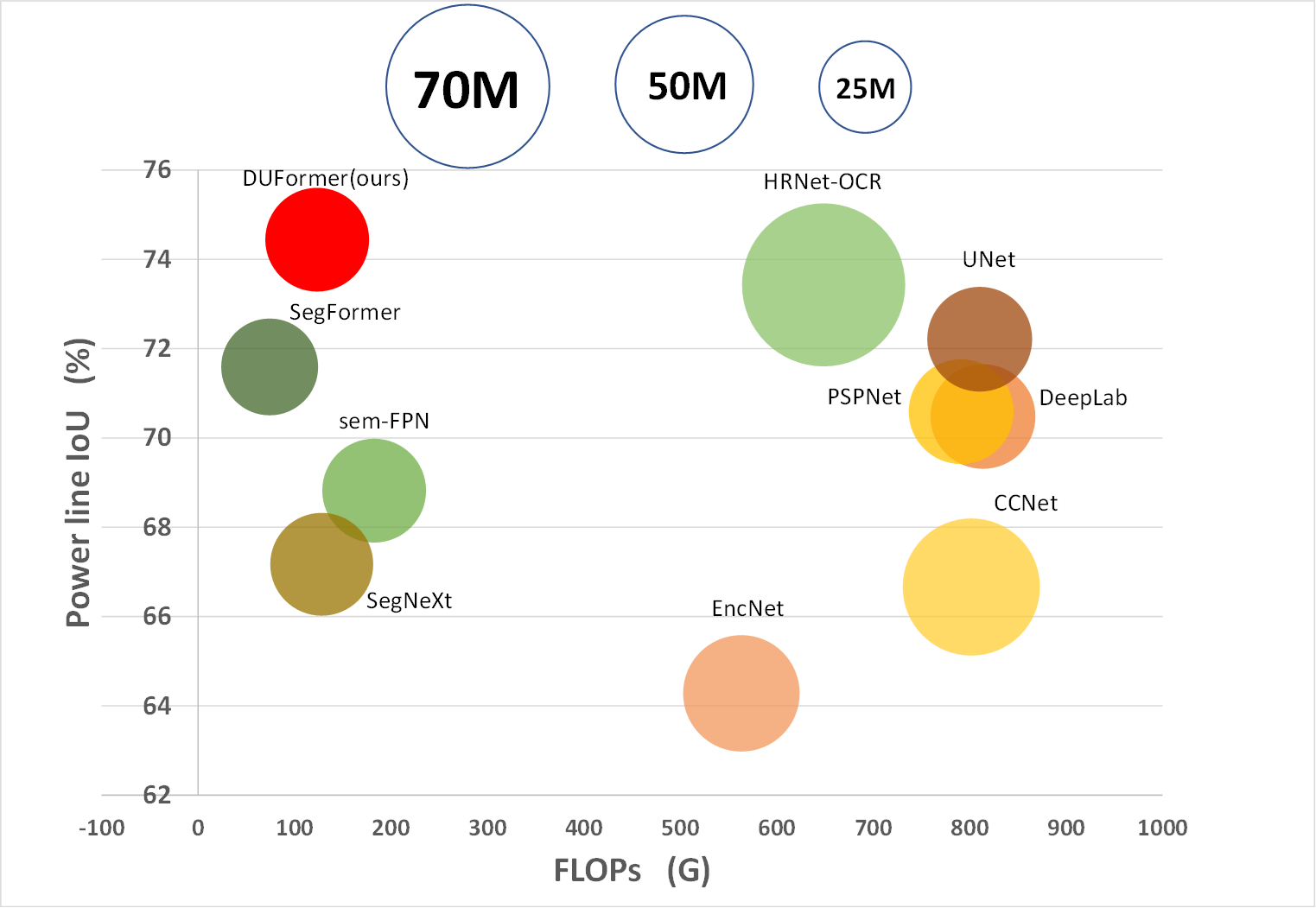}
	\end{center}
\caption{FLOPs and power line IoU for various models on the TTPLA validation set. The FLOPs are calculated at an input resolution of $1024\times1024$.}
	\label{fig:flops}
\end{figure}

Fig.~\ref{fig:flops} presents the visualization results of FLOPs, IoU, and the number of parameters for each model. It is evident that our DUFormer outperforms other models significantly. As the circle moves towards the upper-left corner of the graph, the IoU increases while the FLOPs decrease. In addition, the circle's radius indicates the number of parameters.
\vspace{-1.0em}
\subsection{Visualization of detected power lines}
\begin{figure}
	\begin{center}
		\includegraphics[width=0.9\linewidth]{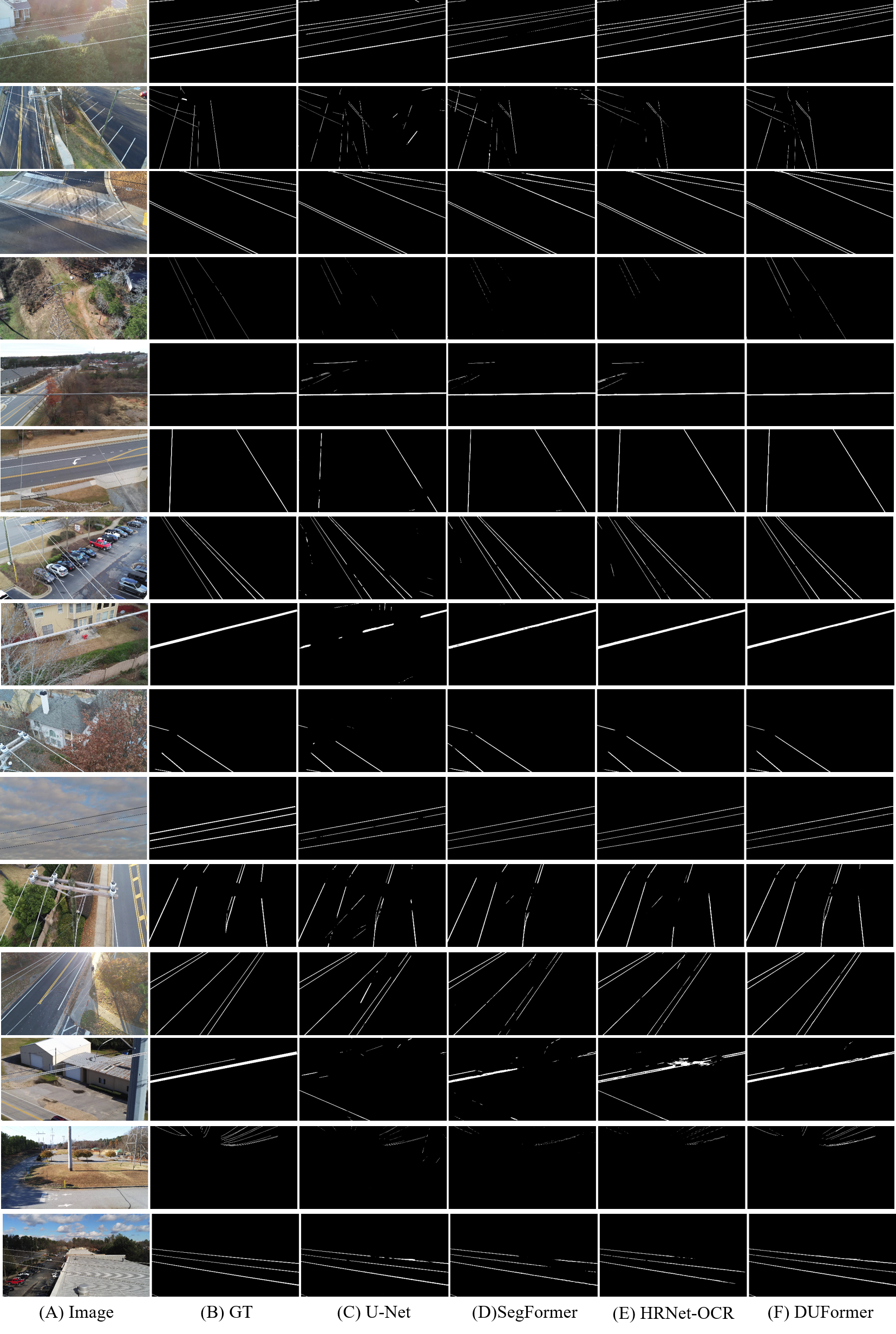}
	\end{center}
\caption{Visualization of detected power lines.}
	\label{fig:comp}
\end{figure}

Fig.~\ref{fig:comp} shows detected power lines in images for different models. It confirms the advantages of our proposed model together with the test accuracy scores.
\section{Other Alabtion Experiments}
	\subsection{Effect of Power Line Aware Block (PLAB)} 
	\noindent To assess the impact of the PLAB on our algorithm, we compare the performance of our algorithm with and without the PLAB. We use DUFormer without the PLAB as the baseline and add the PLAB module to verify its effectiveness. The results in Table~\ref{tal:plab} show that when the PLAB is applied to the first three stages (i.e., shallow layers), the network significantly improves the performance. The IoU metric increases by 1.29\%, and the Precision metric increases by 1.22\%. However, when the PLAB is applied to the network's deep layers, it hurts power line extraction. The Recall metric is even lower than the network's performance without the PLAB by 0.6\%. This suggests that the PLAB module with atrous convolution effectively handles feature maps with much redundant information in shallow layers. In contrast, the PLAB corrupts the strong semantic information of feature maps in deep layers, i.e., stage 4 and stage 5.
	\begin{table}
		\begin{center}
			\caption{Apply PLAB in different stages.}\label{tal:plab}
			\begin{tabular}{c|c|c|c}
				\hline
				PLAB Locations & Precision(\%) & Recall(\%) & IoU(\%) \\
				\hline
				None & 83.13 & 85.91 & 73.15 \\
				Stage1, Stage2, Stage3 & 84.35 & $\bm{86.37}$ & $\bm{74.44}$ \\
				Stage1, Stage2, ..., Stage5 & $\bm{84.38}$ & 85.31 & 73.68 \\	
				\hline
			\end{tabular}
		\end{center}
	\end{table}
	\subsection{Effect of BiscSE module} 
	\noindent The proposed module BiscSE comprises two key elements to enhance the dichotomous segmentation of power lines: the improved channel SE (Bi-channel SE) and the spatial SE (Bi-spatial SE). We evaluate their performance on DUFormer using the standard scSE module as a reference. As shown in Table~\ref{bicsse}, our results indicate that Bi-channel SE improves the Precision metric, while Bi-spatial SE considerably enhances the Recall metric. Therefore, we propose the BiscSE module by integrating the advantages of these two improved methods. Our experiments show that the proposed BiscSE outperforms the original scSE module in the task of power line detection of aerial images.
	\begin{table}
		\begin{center}
			\caption{Results of BiscSE.}\label{bicsse}
			\begin{tabular}{c|c|c|c}
				\hline
				Method &Precision(\%) & Recall(\%) & IoU(\%) \\
				\hline
				Origin. scSE & 84.48 & 85.45 & 73.85 \\
				Bi-channel SE & $\bm{84.92}$ & 85.33 & 74.11 \\
				Bi-spatial SE & 83.45 & 86.04 & 73.5 \\
				BiscSE (ours) & 84.35 & $\bm{86.37}$ & $\bm{74.44}$ \\
				\hline
			\end{tabular}
		\end{center}
	\end{table}